\documentclass{article}

\usepackage{xcolor}
\definecolor{mydarkblue}{rgb}{0,0.15,0.7}
\usepackage[colorlinks, urlcolor=mydarkblue, citecolor=mydarkblue, linkcolor=mydarkblue]{hyperref} 
\newcommand{\vv}[1]{\boldsymbol{#1}}

\usepackage[final]{midl_2018}

\usepackage[utf8]{inputenc} 
\usepackage[T1]{fontenc}    
\usepackage{hyperref}       
\usepackage{url}            
\usepackage{booktabs}       
\usepackage{amsfonts}       
\usepackage{nicefrac}       
\usepackage{microtype}      
\usepackage{amsmath}
\usepackage{algorithmic}
\usepackage{algorithm}
\usepackage{graphicx}
\usepackage{subcaption}
\usepackage{siunitx}
\usepackage{textcomp}

\title{Cancer Metastasis Detection With \\ Neural Conditional Random Field}

\author{
  Yi Li \\
  Baidu Silicon Valley Artificial Intelligence Lab \\
  1195 Bordeaux Dr. Sunnyvale, CA 94089 \\
  \texttt{liyi17@baidu.com} \\
  \And
  Wei Ping \\
  Baidu Silicon Valley Artificial Intelligence Lab \\
  1195 Bordeaux Dr. Sunnyvale, CA 94089 \\
  \texttt{pingwei01@baidu.com} \\
}

\begin{document}

\maketitle

\begin{abstract}
Breast cancer diagnosis often requires accurate detection of metastasis in lymph
nodes through Whole-slide Images (WSIs). Recent advances in deep convolutional
neural networks (CNNs) have shown significant successes in medical image analysis
and particularly in computational histopathology. Because of the outrageous large
size of WSIs, most of the methods divide one slide into lots of small image patches
and perform classification on each patch independently. However, neighboring patches
often share spatial correlations, and ignoring these spatial correlations may
result in inconsistent predictions. In this paper, we propose a neural conditional
random field (NCRF) deep learning framework to detect cancer metastasis in WSIs.
NCRF considers the spatial correlations between neighboring patches through a fully
connected CRF which is directly incorporated on top of a CNN feature extractor.
The whole deep network can be trained end-to-end with standard back-propagation
algorithm with minor computational overhead from the CRF component. The CNN feature
extractor can also benefit from considering spatial correlations via the CRF component.
Compared to the baseline method without considering spatial correlations, we show that
the proposed NCRF framework obtains probability maps of patch predictions with
better visual quality. We also demonstrate that our method outperforms the baseline in cancer
metastasis detection on the Camelyon16 dataset and achieves an average FROC score
of 0.8096 on the test set. NCRF is open sourced at \url{https://github.com/baidu-research/NCRF}.

\end{abstract}

\section{Introduction}

Breast cancer is one of the leading causes of death among women in the United
States \citep{us2017united}. Early cancer diagnosis and treatment play a
crucial role in improving patients’ survival rate \citep{saadatmand2015influence}.
One of the most important early diagnosis is to detect metastasis in lymph nodes
through microscopic examination of hematoxylin and eosin (H\&E) stained
histopathology slides. In recent years, pathologists have been using
Whole-slide Images (WSIs) to distinguish between normal and tumor cells and
localize malignant lesions~\citep{araujo2017classification}. However, manually
detecting tumor cells within extremely large WSIs (e.g., 100,000 $\times$ 200,000 pixels)
can be tedious and time-consuming. Significant discordance on detection results
among different pathologists has also been reported~\citep{elmore2015diagnostic},
where the overall concordance rate of participating pathologists was $75.3\%$.
Therefore, various computer-aided diagnosis (CAD) systems have been developed to
assist pathologists to detect cancer metastasis in WSIs~\citep{brem2005evaluation, apou2014efficient}.

In recent years, deep convolutional neural networks (CNNs) have shown significant 
improvements on a wide range of computer vision tasks on natural images, e.g.
image classification \citep{krizhevsky2012imagenet, simonyan2014very, he2016deep},
object detection \citep{girshick2014rich, girshick2015fast}, and semantic segmentation
\citep{long2015fully}. Similarly, a few promising studies have also applied deep CNNs
to analyse medical images and particularly WSIs
\citep{wang2016deep, geccer2016detection, hou2016patch, liu2017detecting, vang2018deep, zhu2017deep, zhu2018adversarial},
among which \citep{wang2016deep} won the Camelyon16 challenge \citep{camelyon16}
for metastasis detection. Because of the extremely large size of WSIs,
most of the studies first extracted small patches (e.g. 256 $\times$ 256 pixels)
from WSIs, and trained a deep CNN to classify these small patches into normal
or tumor regions. A probability map of the original WSI being tumor or normal
at patch level was later obtained and metastasis detection was performed based
on this probability map. However, the small patches and their neighbors often
share spatial correlations. Because the patches were extracted and trained
independently, the spatial correlations were not modeled explicitly. Therefore,
during inference time, the predictions over neighboring patches may be inconsistent,
and the patch level probability map may contain isolated outliers~\citep{kong2017cancer, zanjani2018cancer}.

To explicitly model the spatial correlations between neighboring patches, Kong et al.~\citep{kong2017cancer}
recently proposed Spatio-Net that uses 2D Long Short-Term Memory (LSTM) layers
to capture the spatial correlations based on patch features extracted from a CNN
classifier. However, Spatio-Net uses a two-stage training approach, and therefore
the CNN feature extractor is not aware of the spatial correlations~\citep{zanjani2018cancer}.
In parallel to our work and very recently, we notice a similar work~\citep{zanjani2018cancer}
that also uses features extracted from a CNN classifier to represent neighboring
patches. Conditional random field (CRF) is then applied on these patch features
to model spatial correlations and refine the predicted probability map during
a post-processing stage. In addition to the same issue due to the two stage framework,
there is a significant computational overhead during the CRF post-processing, and
the authors have to select a limited number of features (e.g., 5 reported in~\citep{zanjani2018cancer})
from the original high dimensional patch representations
to perform the CRF inference algorithm on CPU.

In this paper, we propose an alternative method for modeling spatial correlations
between neighboring patches through neural conditional random field (NCRF).
NCRF is a probabilistic graphical model that combines both neural networks and
conditional random fields. It has been used for sequence labeling~\citep{artieres2010neural, peng2009conditional}
and semantic image segmentation~\citep{chen2018deeplab}. We implemented NCRF based
on the idea of conditional random fields as recurrent neural networks~\citep{zheng2015conditional}
that directly incorporates a fully connected CRF on top of the CNN feature extractor.
The marginal label distribution of each patch is obtained through the mean-field approximate
inference algorithm. The whole deep network can be trained in an end-to-end manner with the
standard back-propagation algorithm, avoiding the post-processing stage. Because
the mean-field inference algorithm is also performed on GPU, the CRF component
introduces minor computational overhead and allows very large feature dimensions,
e.g. 512 from the ResNet architecture~\citep{he2016deep}. The CNN feature
extractor also benefits from jointly training with the CRF component, since it
is now aware of the spatial correlations between neighboring patches. Compared
to the baseline method that does not consider patch spatial correlations, we show
that, 1) NCRF improves the visual quality of the probability map, 2) NCRF improves
the CNN feature extractor, and 3) NCRF improves the performance of cancer metastasis
detection. On the test set of the Camelyon16 challenge, the best average
free response receiver operating characteristic (FROC) score of NCRF is 0.8096,
which outperforms the previous best average FROC score of $0.8074$ reported in~\citep{wang2016deep}.

\section{Method}

\begin{figure}[h] 
  \centering
  \includegraphics[width=0.99\textwidth]{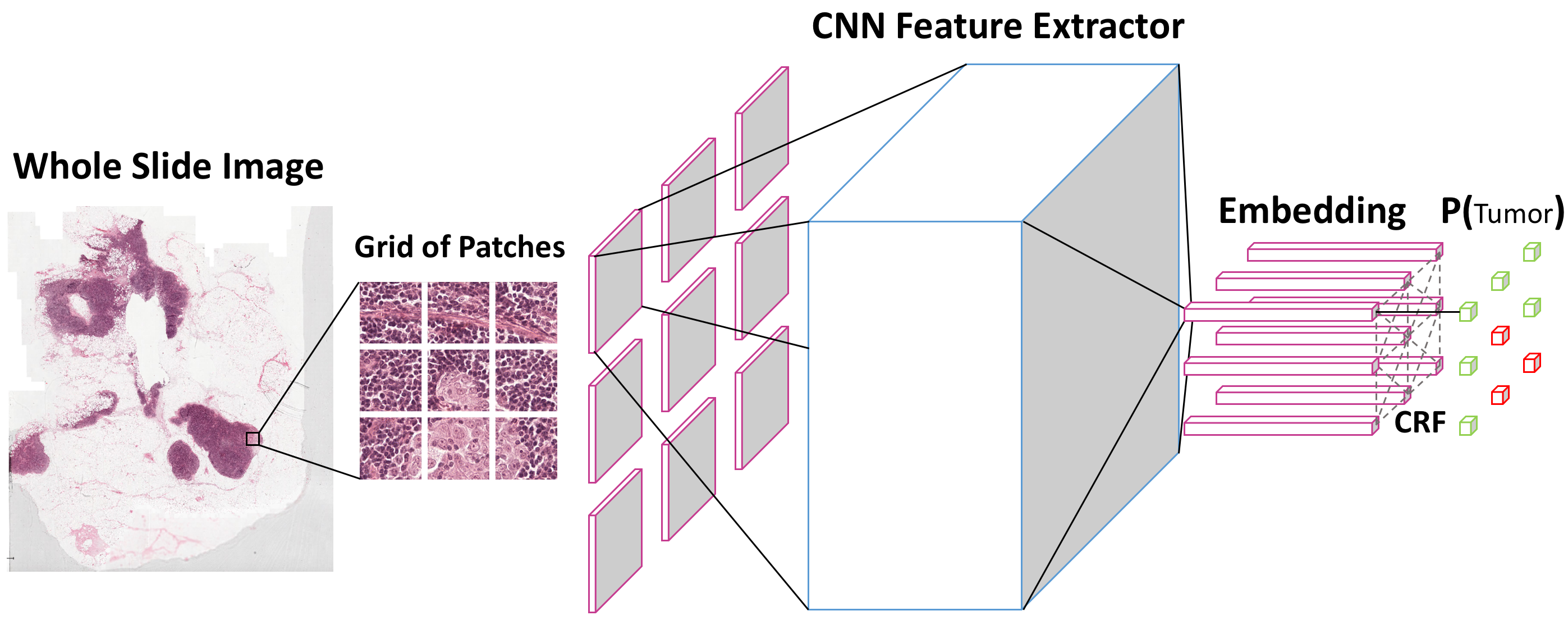}  
  \caption{The architecture of NCRF model.}\label{fig1}
\end{figure}
In this section, we describe the details of the proposed neural conditional
random field~(NCRF) model. Figure~\ref{fig1} shows the overall architecture of NCRF.
It has two major components: CNN and CRF. The CNN component acts as a feature extractor,
that takes a grid of patches as input, and encodes each patch as a fixed-length
vector representation (i.e. embedding). The CRF component takes the grid of embeddings
as input and models their spatial correlations. The final output from the CRF
component is the marginal distribution of each patch being normal or tumor given
the grid of patch embeddings. We illustrate the details of each component in the
next two sections.

\subsection{Patch Embedding With CNN}
To extract comprehensive feature representation of each patch, we employ two ResNet
architectures~\citep{he2016deep} that have proven to be powerful in image classification
task, ResNet-18 and ResNet-34. For each architecture, we use the activations
after the average pooling layer as the embedding for each patch. The embedding
size is 512 for both ResNet-18 and ResNet-34, which is much larger than the
embedding size of 5 reported in~\citep{zanjani2018cancer} after feature selection.

\subsection{Spatial Modeling With CRF}
In this section, we describe the methodology details of the CRF component. We
denote a grid of patch embeddings obtained from CNN as $\vv x = \{x_i\}_{i=1}^N$,
where $N$ is the number of patches within the grid, e.g. 25 for a grid of 5 $\times$ 5.
Let $\mathbf{Y} = \{y_i\}_{i=1}^N$ be the random variables associated with each
patch $i$, that represents the label of patch $i$ takes a value from $\{normal,\ tumor\}$.
The conditional distribution $P(\mathbf{Y} \mid \vv x)$ can be modeled as a CRF
with a Gibbs distribution of
\begin{eqnarray}
P(\mathbf{Y} = \vv y \mid \vv x) = \frac{1}{Z(\mathbf{x})} \exp(-E(\vv y, \vv x))
\end{eqnarray}
where $E(\vv y, \vv x)$ is the energy function that measures the cost of $\mathbf{Y}$
taking a specific configuration $\vv y$ given $\vv x$, and $Z(\vv x)$
is the partition function that insures $P(\mathbf{Y} = \vv y \mid \vv x)$ is a valid
probability distribution. In a fully-connected pairwise CRF \citep{krahenbuhl2011efficient},
the energy function is given by:
\begin{eqnarray}
E(\mathbf{y}, \vv x) = \sum_i \psi_u (y_i) + \sum_{i<j} \psi_p(y_i, y_j)
\end{eqnarray}
where $i, j$ ranges from 1 to $N$. $\psi_u (y_i)$ is the unary potential that measures
the cost of patch $i$ taking the label $y_i$ given the patch embedding $x_i$, and
$\psi_p(y_i, y_j)$ is the pairwise potential that measures the cost of jointly
assigning patch $i, j$ with label $y_i, y_j$ given the patch embeddings $x_i, x_j$.
Pairwise potential $\psi_p(y_i, y_j)$ models spatial correlations
between neighboring patches, and would encourage low cost for assigning $y_i, y_j$
with the same label if $x_i, x_j$ are similar. We implement the unary potential
$\psi_u (y_i)$ as the negative log-likelihood of patch $i$ taking label $y_i$,
which is the negative logit for label $y_i$ before the softmax layer of the CNN classifier.
We implement the pairwise potential as the weighted cosine distance between $x_i, x_j$:
\begin{eqnarray}
\psi_p(y_i, y_j) = \mathbb{I}(y_i = y_j) \cdot w_{i,j} \left(1 - \frac{x_i \cdot x_j}{\left\Vert x_i \right\Vert \left\Vert x_j \right\Vert} \right)
\label{eq1}
\end{eqnarray}
where $\mathbb{I}(y_i = y_j)$ is the indicator function that checks the label compatibility
between $y_i, y_j$, and $w_{i,j}$ is a trainable weight which controls the correlation
strength between two patches $i, j$ within the grid. Typically, fully connected CRF
also includes another distance term for pairwise potential that encodes the spatial
distance between two patches $i, j$~\citep{krahenbuhl2011efficient}. However we
did not observe clear improvements by including such distance term, and if we put a trainable
coefficient before the term, the coefficient was pushed to zero during training. On the
other hand, we observed the trainable weight $w_{i,j}$ correlated well with the
relative distances between different patches within the grid after model converged, and we show
this in the results section.

In order to train the CNN-CRF architecture end-to-end with the standard back-propagation
algorithm, we need to obtain the marginal distribution of each patch label $y_i$,
so that it can be used to compute the cross-entropy loss with respect to the ground
truth labels~\citep{zheng2015conditional}. However, exact marginal inference is
intractable, and we use mean-field approximate inference, where the original CRF
distribution $P(\mathbf{Y})$\footnote{We omit the dependency on $\vv x$ here for clarity.} 
is approximated with a simpler distribution $Q(\mathbf{Y})$,
that can be written as the product of marginal distributions of each individual
patch $i$, $Q(\mathbf{Y}) = \prod_i^N Q_i(y_i)$. By minimizing the KL divergence between
$Q(\mathbf{Y})$ and $P(\mathbf{Y})$, $\mathbb{KL}(Q(\mathbf{Y}) \| P(\mathbf{Y}))$,
we derive the update step for each marginal distribution $Q_i(y_i)$~\citep{murphy2012machine}:
\begin{eqnarray}
\log Q_i(y_i) = \mathbb{E}_{-Q_i} \left[ \log \tilde{P}(\mathbf{Y}) \right] + \text{const}
\end{eqnarray}
where $\mathbb{E}_{-Q_i} \left[ f(\mathbf{Y}) \right]$ means taking the expectation of $f(\mathbf{Y})$
with respect to all the variables except $y_i$, and $\tilde{P}(\mathbf{Y}) = \exp(-E(\vv y, \vv x)) $ is
the unnormalized CRF distribution. The mean-field inference algorithm is summarized
in Algorithm~\ref{algo1}

\begin{algorithm} 
\caption{Mean-field inference algorithm}
\begin{algorithmic}
\label{algo1}
\STATE compute $\psi_u (y_i)$ for all $i$ and  $\psi_p (y_i, y_j)$ for all $i,j$
\STATE $\log \tilde{P}(\mathbf{Y}) \gets -\left[ \sum_i \psi_u (y_i) + \sum_{i<j} \psi_p(y_i, y_j) \right]$
\STATE initialize $Q_i(y_i) \gets \exp(-\psi_u (y_i))$ for all $i$
\vspace{0.15em}
\STATE normalize $Q_i(y_i)$ for all $i$
\vspace{0.15em}
\FOR{T iterations}
\vspace{0.15em}
\STATE $\log Q_i(y_i) \gets \mathbb{E}_{-Q_i} \left[ \log \tilde{P}(\mathbf{Y}) \right]$ for all $i$
\STATE normalize $Q_i(y_i)$ for all $i$
\ENDFOR
\end{algorithmic}
\end{algorithm}
Finally, after a fixed number of mean-field iterations, we use the approximate
marginal distribution of each patch label $Q_i(y_i)$ to compute the cross-entropy
loss and train the whole model with back-propagation algorithm.

\section{Experiments}
In this section, we present empirical evaluations on the proposed NCRF
method. We demonstrate its advantages over the baseline method without CRF in
three aspects: 1) NCRF obtains smoother probability maps with sharp boundaries
than the baseline method, 2) NCRF achieves higher patch level classification
accuracies from the CNN feature extractor than the baseline method, and 3) NCRF
outperforms the baseline method in cancer metastasis detection.

\subsection{Data Preparation}
We conducted all the experiments based on the Camelyon16 dataset~\citep{camelyon16, bejnordi2017diagnostic}.
This dataset includes 160 normal and 110 tumor WSIs for training, 81 normal and
49 tumor WSIs for testing. Slides were exhaustively annotated by pathologists
in pixel level, with a few exceptions reported in~\citep{liu2017detecting}.
We conducted all the experiments on $40 \times$ magnification. We used the Otsu
algorithm~\citep{otsu1979threshold, wang2016deep} to exclude the background regions
of each training slide. We used Normal\_001 to Normal\_140 and Tumor\_001 to Tumor\_100
for training, and the rest of training slides for validation. To generate patches,
we first randomly picked one slide and then randomly sampled a coordinate from the
slide as the center of the patch. We randomly sampled $200,000$~ $768\times768$
patches from the tumor regions of the tumor slides as positive samples. We randomly
sampled $200,000$~ $768\times768$ patches from the non-tumor non-background regions
of the tumor slides and the non-background regions of the normal slides as negative
samples. Hard negative mining~\citep{wang2016deep} was also applied to sample
more patches from the tissue boundary regions.

\subsection{Implementation Details}
NCRF was implemented with PyTorch-0.3.1~\citep{paszke2017automatic} and trained with
NVIDIA GeForce GTX 1080 Ti GPU. The mean-field inference algorithm for the CRF
component was performed 10 iterations for all the architectures. The CRF component 
introduces less than 0.1 seconds computational overhead per batch iteration,
since the mean-field inference algorithm is also performed on GPU. During the
training stage, a batch size of 20 $768 \times 768$ patches were feed into the model. Each
$768 \times 768$ patch was further split into a $3 \times 3$ grid of $256 \times 256$
patches and their corresponding labels were retrieved. The pixel values of patches
were normalized by subtracting $128$ and dividing $128$. During training, color
jitter was added using torchvision tranforms with parameters adopted from~\citep{liu2017detecting}
:brightness with a maximum delta of 64/255, contrast with a maximum delta of 0.75,
saturation with a maximum delta of 0.25, and hue with a maximum delta of 0.04.
Patches were also randomly flipped and rotated with multiplies of $\ang{90}$.
We used stochastic gradient descent of learning rate $0.001$ and a momentum of
$0.9$ to optimize all the architectures for 20 epochs. Each architecture was
repetitively trained 5 times with different random seeds for parameters initialization.
During inference time, probability maps were generated with a stride of 64 (level 6).

\subsection{NCRF Obtains Smooth Probability Maps}

\begin{figure}
    \centering
    \begin{subfigure}[b]{0.31\textwidth}
        \includegraphics[width=\textwidth]{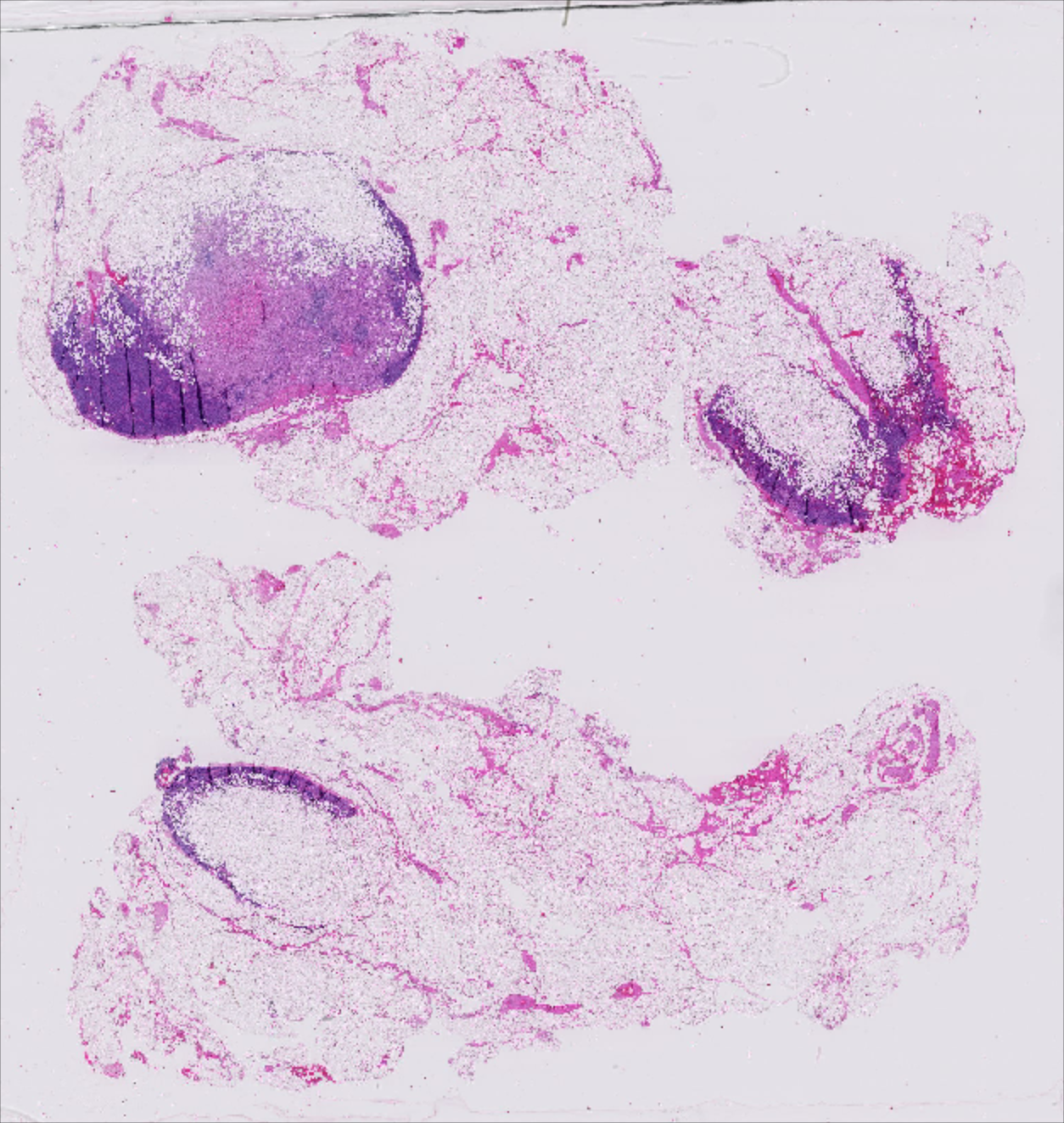}
        \caption{}
        \label{fig2:a}
    \end{subfigure}
    ~ 
    \begin{subfigure}[b]{0.31\textwidth}
        \includegraphics[width=\textwidth]{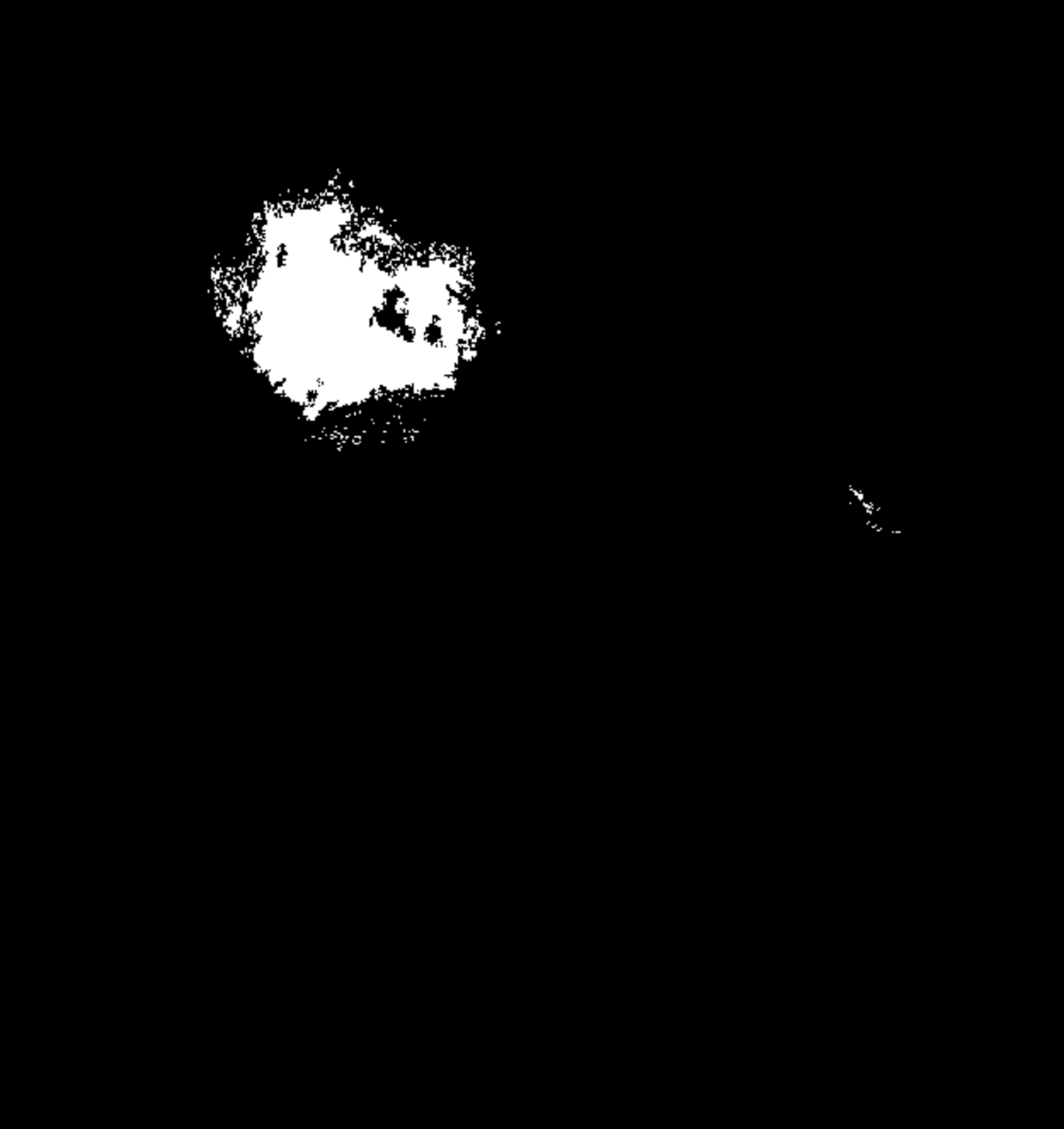}
        \caption{}
        \label{fig2:b}
    \end{subfigure}

    ~ 
    \begin{subfigure}[b]{0.31\textwidth}
        \includegraphics[width=\textwidth]{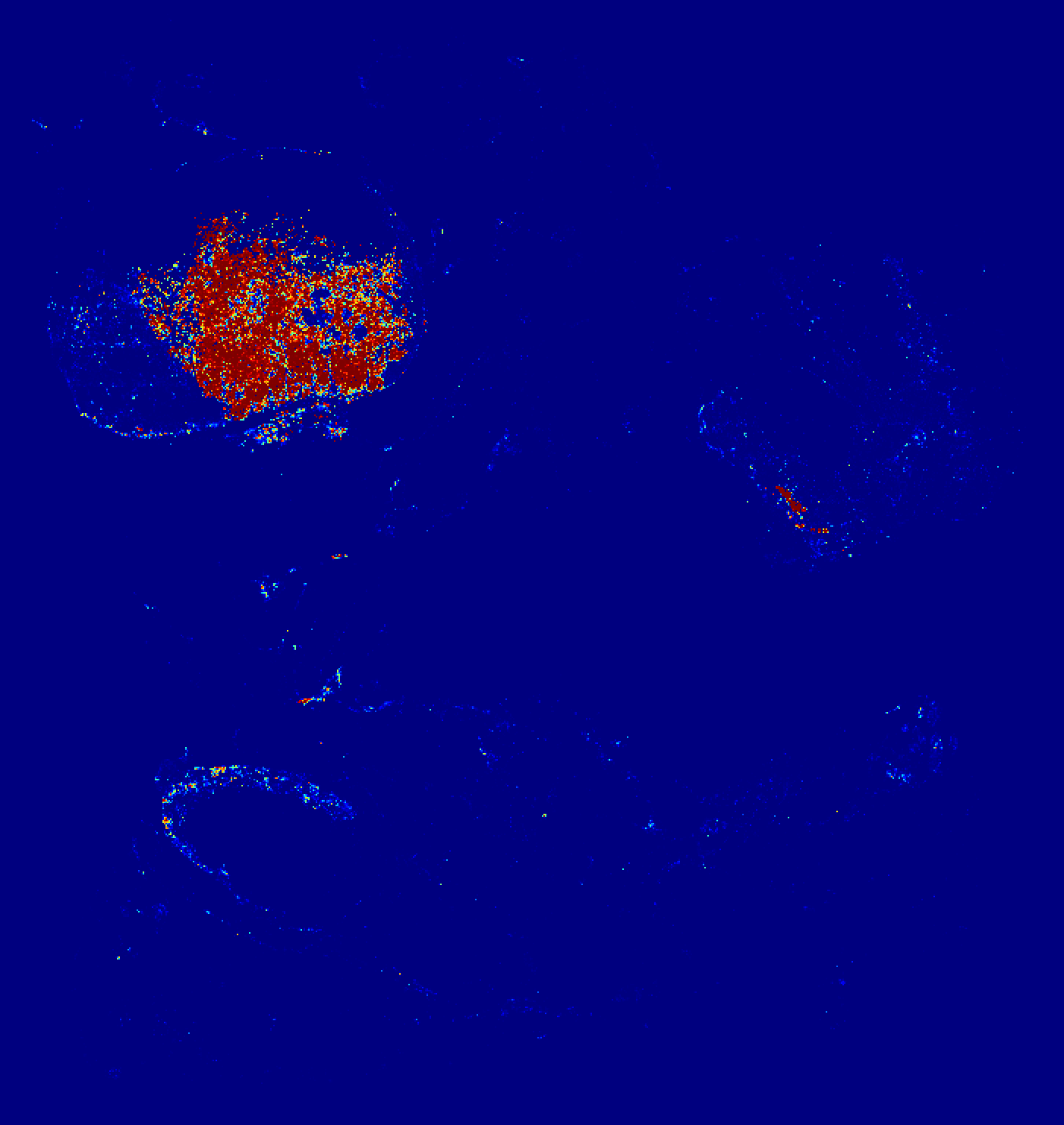}
        \caption{}
        \label{fig2:c}
    \end{subfigure}
    ~ 
    \begin{subfigure}[b]{0.31\textwidth}
        \includegraphics[width=\textwidth]{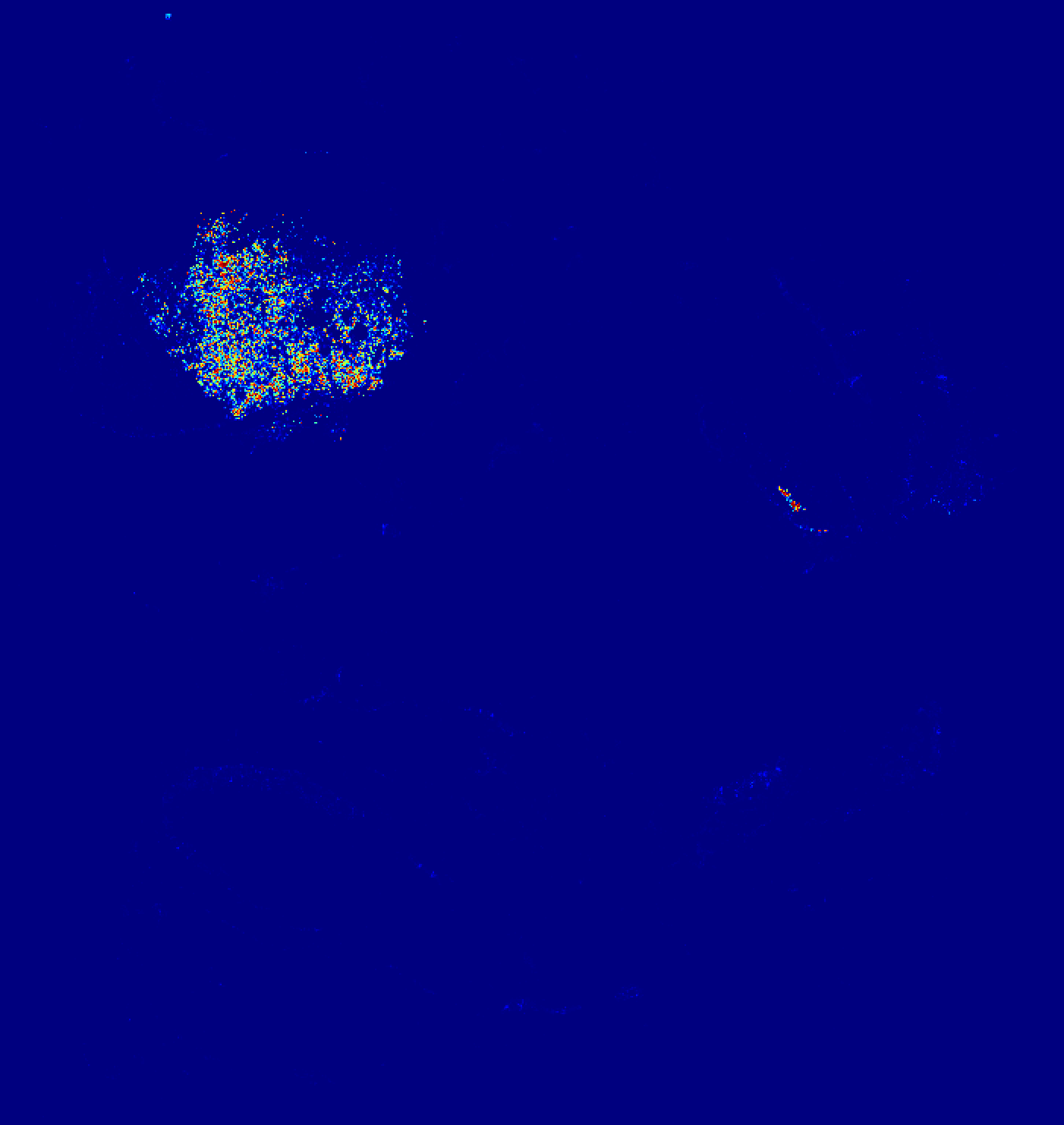}
        \caption{}
        \label{fig2:d}
    \end{subfigure}
    ~ 
    \begin{subfigure}[b]{0.31\textwidth}
        \includegraphics[width=\textwidth]{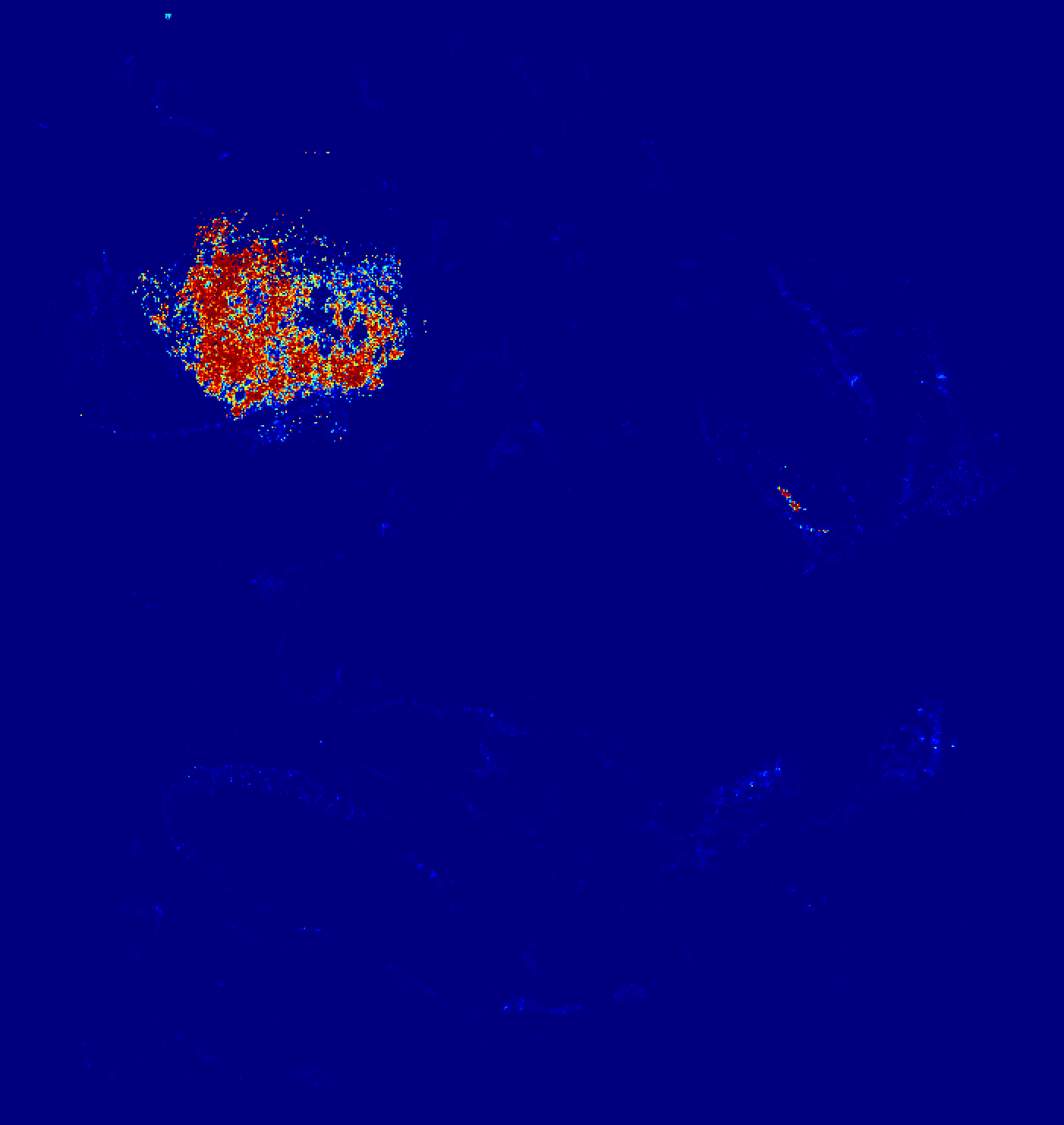}
        \caption{}
        \label{fig2:e}
    \end{subfigure}
    ~ 
    \caption{Predicted probability maps of Test\_026 from ResNet-18 with random seed 0.
    (a) original WSI, (b) ground truth annotation, (c) baseline method, (d) baseline method
    with hard negative mining, (e) NCRF with hard negative mining}\label{fig2}
\end{figure}

Figure~\ref{fig2} shows the predicted probability maps of Test\_026 from the
baseline method, baseline method with hard negative mining, and NCRF with hard
negative mining, all based on the ResNet-18 architecture with random seed 0.
We can see the probability map from the baseline method that does not consider
spatial correlations tends to contain isolated outlier predictions, which
significantly increases the number of false positives. Hard negative mining
significantly reduces the number of false positives for the baseline method,
but the probability density among the ground truth tumor regions is also decreased,
which decreases model sensitivity. Compared to the baseline method with hard
negative mining, NCRF with hard negative mining not only achieves low false positives
but also maintains high probability density among the ground truth tumor regions
with sharp boundaries. In fact, NCRF detects two more tumor regions in Test\_026
compared to the baseline method in this case.

\subsection{NCRF Improves CNN Feature Extractor}
\begin{figure}[h] 
  \centering
  \includegraphics[width=0.99\textwidth]{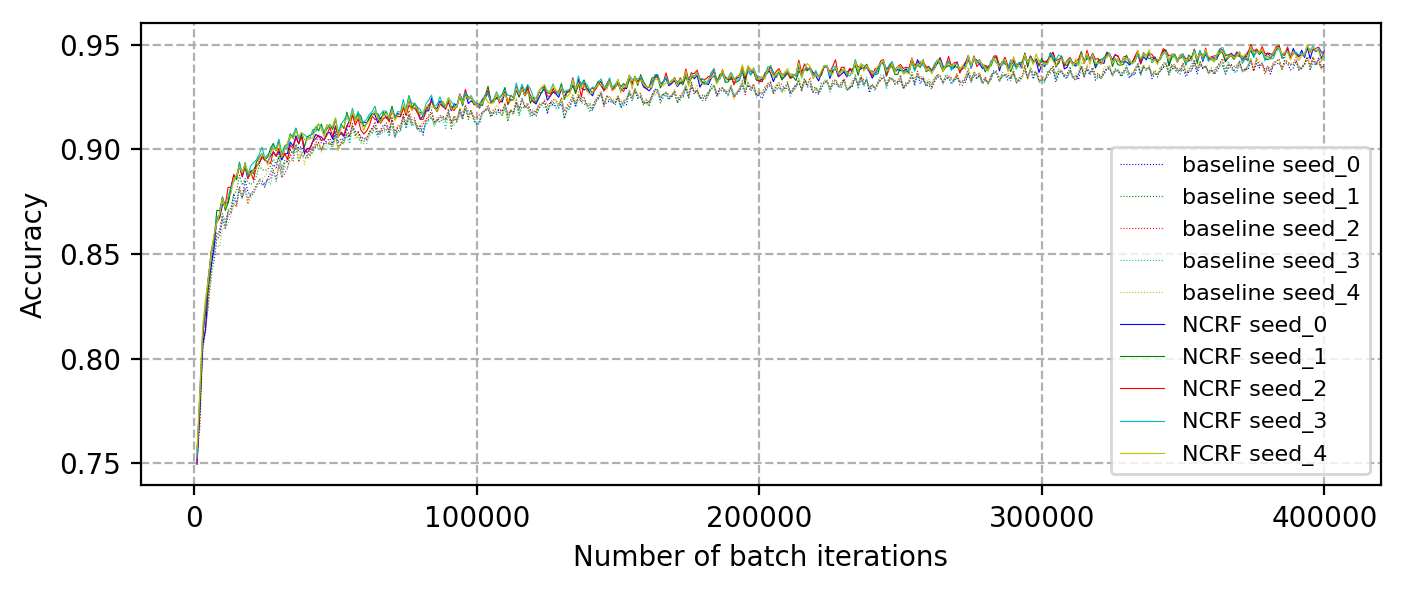}  
  \caption{Patch classification accuracies of the baseline method and NCRF based
   on the ResNet-18 architecture throughout training with different random seeds
   for initialization, best view in color.}\label{fig3}
\end{figure}

\begin{figure}[h] 
  \centering
  \includegraphics[width=0.6\textwidth]{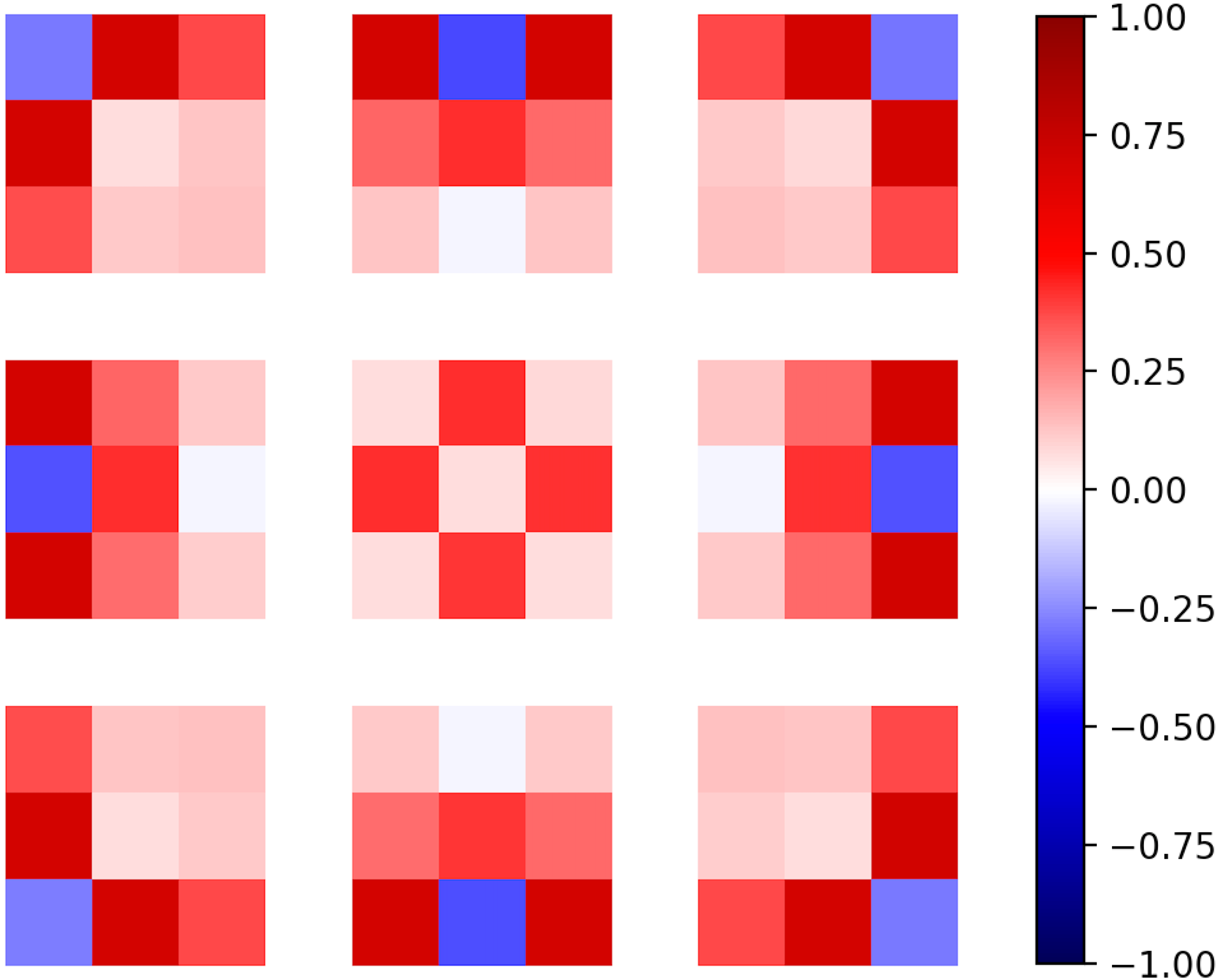}  
  \caption{Visualization of the learned weights for pairwise potential defined
  in Equation~\ref{eq1} arranged according to the position of each patch within
  the $3 \times 3$ grid. The position of each sub-figure within the whole figure,
  represents the position of each patch within the $3 \times 3$ grid. The color
  map of each sub-figure shows the pairwise weights of all the patches with respect
  to the patch that the sub-figure represents. Weights were extracted from ResNet-18
  with random seed 0.}\label{fig4}
\end{figure}

\begin{table}[t]
  \caption{Patch classification accuracies on the validation set.}
  \label{sample-table}
  \centering
  \begin{tabular}{lll}
    \toprule
    \cmidrule{1-3}
                                 & baseline              & NCRF               \\
    \midrule
    ResNet-18\citep{he2016deep}  & 0.9242 $\pm$ 0.0007   & 0.9296 $\pm$ 0.0013\\
    ResNet-34\citep{he2016deep}  & 0.9251 $\pm$ 0.0007   & 0.9338 $\pm$ 0.0014\\
    \bottomrule\label{tab1}
  \end{tabular}
\end{table}

NCRF improves the CNN feature extractor by incorporating spatial correlations
between neighboring patches during training. Figure~\ref{fig3} shows the patch
classification accuracies of the baseline method and NCRF based on the ResNet-18
architecture throughout training with different random seeds for initialization.
NCRF consistently achieves higher training accuracies than the baseline method
across different random seeds for initialization. Table~\ref{tab1} shows the
best patch classification accuracies of the baseline method and NCRF on the
validation set across different random seeds. NCRF consistently improves the patch
classification accuracies on both ResNet-18 and ResNet-34, compared to the baseline
method. These results show the CNN feature extractor benefits from the end-to-end
joint training with CRF, compared to the previous two-stage training framework
proposed in \citep{kong2017cancer, zanjani2018cancer}, where the CNN feature extractor
is not aware of the patch spatial correlations.

The learned weights for pairwise potential defined in Equation~\ref{eq1} also show
strong spatial pattern. Figure~\ref{fig4} shows the visualization of the learned
weights arranged according to the position of each patch within the $3 \times 3$ grid.
For example, the sub-figure in the center represents the center patch within the
grid, and its color map shows the pairwise weights of all the patches with respect
to the center patch. We can see that for a specific patch within the grid, its
closest neighboring patches have the largest pairwise weights, suggesting the
label distribution of each patch is strongly correlated with their closest neighbors
within the grid.

\subsection{NCRF Improves Cancer Metastasis Detection}

\begin{figure}[h] 
  \centering
  \includegraphics[width=0.7\textwidth]{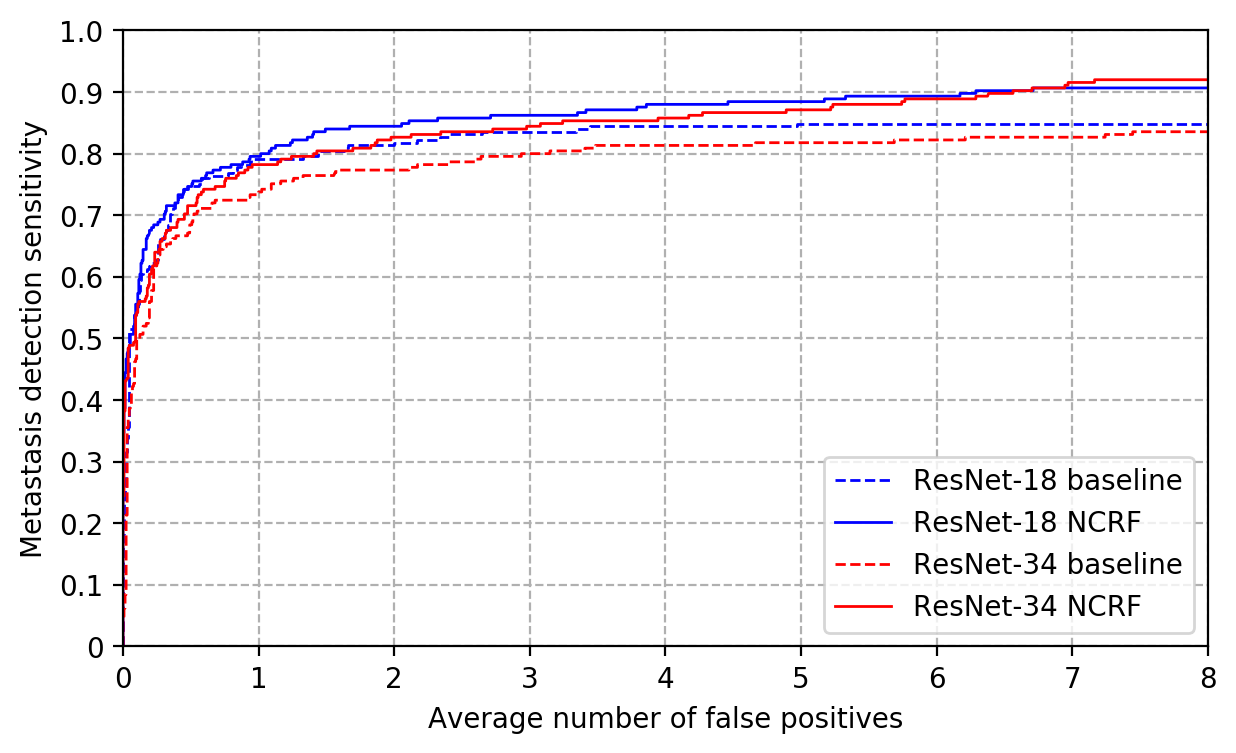}  
  \caption{FROC curves of the baseline method and NCRF on the test set based on
  ResNet-18 and ResNet-34 with random seed 0.}\label{fig5}
\end{figure}

\begin{table}[t]
  \caption{Average FROC score on the test set.}
  \label{sample-table}
  \centering
  \begin{tabular}{lll}
    \toprule
    \cmidrule{1-3}
                                 & baseline              & NCRF               \\
    \midrule
    ResNet-18\citep{he2016deep}  & 0.7825 $\pm$ 0.0102   & 0.7934 $\pm$ 0.0168\\
    ResNet-34\citep{he2016deep}  & 0.7444 $\pm$ 0.0121   & 0.7704 $\pm$ 0.0171\\
    \bottomrule\label{tab2}
  \end{tabular}
\end{table}

We evaluated the performance of cancer metastasis detection of NCRF based on the
average free-response receiver operating characteristic (FROC) score on the Camelyon16
test set. Given a list of predicted coordinates of cancer metastasis, the average
FROC score is defined as the average detection sensitivity at 6 predefined false-positive
rates per slide: $1/4$, $1/2$, $1$, $2$, $4$ and $8$. Higher average FROC score
means better detection performance. We used the non maximum suppression algorithm~\citep{liu2017detecting},
to obtain the coordinates of cancer metastasis based on a given probability map.

Figure~\ref{fig5} shows the curves of FROC scores of the baseline method and NCRF on
the test set based on ResNet-18 and ResNet-34 with random seed 0. Table~\ref{tab2}
shows the average FROC scores of the baseline method and NCRF on the test set based on
ResNet-18 and ResNet-34 across different random seeds. NCRF also consistently
improves the FROC scores on both architectures, compared to the baseline method.
The best average FROC score from NCRF is $0.8096$ based on ResNet-18 with random
seed 0, which outperforms the previous best average FROC score of $0.8074$ reported
in~\citep{wang2016deep}.

\section{Discussion}
In this paper, we propose a neural conditional random field (NCRF) 
framework to detect cancer metastasis in Whole-slide Images (WSIs). NCRF is
able to consider the spatial correlations between neighboring patches through
the fully connected CRF component. Compared to previous methods, the CRF component
is unified with the CNN feature extractor, and the whole model can
be trained end-to-end with standard back-propagation algorithm. Because of this
joint training framework, the CNN feature extractor also benefits from considering
the spatial correlations while the CRF component introduces minor computational
overhead. Compared to the baseline method without considering patch spatial
correlations, NCRF obtains not only smoother probability maps but also better
performances in cancer metastasis detection. NCRF is also a general technique that can
be applied to other settings in pathology analysis, e.g. multiple-instance learning~\citep{kandemir2014digital},
when only whole slide level annotation is available. One future direction is using a
grid of more than $3 \times 3$ patches as input, since it corresponds to
a larger receptive field and may achieve better performance in cancer metastasis
detection. Another interesting future direction is to compare NCRF with the baseline
method that also uses patch coordinates as input, since patch localization is explicitly
modeled in this case.


\small

\bibliographystyle{abbrv}
\bibliography{document} 

\end{document}